\documentclass[conference]{IEEEtran}
\IEEEoverridecommandlockouts
\usepackage{newtxtext}
\usepackage{float}
\usepackage[numbers]{natbib}
\usepackage{algorithmic}
\usepackage{array}
\usepackage{textcomp}
\usepackage{multirow}
\usepackage{stfloats}
\usepackage[linesnumbered,ruled,vlined]{algorithm2e}
\usepackage{url}
\usepackage{verbatim}
\usepackage{etoolbox}
\usepackage{graphicx}
\usepackage{amsmath,amssymb,amsfonts}
\usepackage{fancyhdr}

\def\tsc#1{\csdef{#1}{\textsc{\lowercase{#1}}\xspace}}
\tsc{WGM}
\tsc{QE}
\tsc{EP}
\tsc{PMS}
\tsc{BEC}
\tsc{DE}

\def\BibTeX{{\rm B\kern-.05em{\sc i\kern-.025em b}\kern-.08em
    T\kern-.1667em\lower.7ex\hbox{E}\kern-.125emX}}
\begin{document}

\pagestyle{fancy}        
\fancyhf{}               
\fancyfoot[C]{\thepage}  
\renewcommand{\headrulewidth}{0pt}
\renewcommand{\footrulewidth}{0pt}
\thispagestyle{fancy}    

\title{Federated Knowledge Distillation for Multi-Model Architectures Lithography Hotspot Detection}

\author{
    \IEEEauthorblockN{
        Yuqi Li$^{1, \dag}$, Xinyou Lin$^{2, \dag}$, Yanli Li$^{3}$, Kai Zhang$^{4}$, Chuanguang Yang$^{5}$, \\ 
        Zhongliang Guo$^{6}$, Jianping Gou$^{7}$, Tingwen Huang$^{4}$, Yingli Tian$^{1, *}$%
        \thanks{$^{\dag}$Yuqi Li and Xingyou Lin are co-first authors of this work and contributed equally. }%
        \thanks{$^{*}$Corresponding author.}
    }
    \IEEEauthorblockA{
        \textit{$^1$City University of New York} \\
        \textit{$^2$Sichuan Agricultural University} \quad \textit{$^3$The University of Sydney} \\
        \textit{$^4$Shenzhen University of Advanced Technology} \quad \textit{$^5$Institute of Computing Technology, Chinese Academy of Sciences} \\
        \textit{$^6$University of St Andrews} \quad \textit{$^7$Southwest University}
    }
}
\maketitle

\begin{abstract}
As a special type of multimedia data, Lithography Hotspot Detection (LHD) training often requires stronger privacy protection than conventional multimedia data, and federated learning provides a promising potential solution to this challenge. However, existing approaches rely solely on either parameter aggregation or Knowledge Distillation (KD), failing to fully exploit the potential of collaborative learning. To address this, we propose \textit{FedKD-hybrid}, a novel framework that synergizes the strengths of both paradigms. Specifically, FedKD-hybrid utilizes a public dataset to facilitate consensus, where clients exchange both parameters of agreed-upon layers and logits. This hybrid information is aggregated to refine local models, enhancing knowledge transfer. Extensive experiments on ICCAD-2012 and real-world FAB datasets demonstrate that FedKD-hybrid consistently outperforms state-of-the-art methods in both effectiveness and robustness.

\end{abstract}


\section{Introduction}
In the information age, integrated circuits underpin modern computing technologies and represent a cornerstone of semiconductor innovation~\cite{liebmann2006reducing,yang2026attribution,jiang2025towards,li2026comprehensive}. As manufacturing scales into deep sub-micron and nanometer regimes, traditional lithography increasingly approaches its physical limits. Insufficient resolution causes growing mismatches between designed photomasks and printed wafer patterns, leading to severe lithography hotspot issues~\cite{pan2023lithography}.

Early lithography hotspot detection (LHD) relied on physics-based simulations to identify potential defects~\cite{2007Automated}. Although accurate, these methods are computationally prohibitive for large-scale use. Pattern matching techniques improved efficiency by applying predefined design rules~\cite{Fan2017Improved}, but sacrificed accuracy. Recently, learning methods have emerged as a more effective solution, achieving higher accuracy and flexibility by learning from historical lithography data~\cite{2021Lithography}.

In practice, LHD is used to considered as special form of multimodal industrial data analysi. Specifically, hotspot detection does not rely solely on layout images, but often involves multiple heterogeneous modalities, such as layout geometry representations, lithography simulation outputs, process variation parameters, and auxiliary design features. Effectively modeling and fusing these heterogeneous modalities is critical for accurate hotspot prediction, yet also increases the complexity of model training and deployment.

Despite its promise, traditional ML usually raises privacy concerns in distributed real-world deployment due to centralized data collection~\cite{li2025threats}. This issue is particularly critical for LHD, where multimodal lithography data are often highly sensitive and proprietary, and cannot be freely shared across foundries or design houses. To mitigate this research gap, federated learning (FL) was introduced by Google in 2016~\cite{2016Communication}. Under the FL paradigm, clients locally train their models and only share model updates, enabling collaborative learning while preserving data privacy.

Currently, two major research pipelines have been explored in FL to facilitate knowledge transfer: parameter-based model aggregation~\cite{pan2023lithography,qi2025towards,qi2025t2veval}
and non-parameter-based knowledge distillation~\cite{arazzi2025secure,li2025ddtime,li2025mmt}. While both approaches demonstrate effectiveness under synchronous/asynchronous and homogeneous/heterogeneous model settings, they typically transfer only partial information, such as model parameters or intermediate knowledge representations (e.g., logits), which may be insufficient for fully capturing the rich multimodal characteristics of LHD data.

In this study, we address the lithography hotspot detection (LHD) task and propose a novel framework termed federated hybrid knowledge distillation (FedKD-hybrid) to achieve richer knowledge transfer and enhanced model performance in federated settings. Specifically, FedKD-hybrid establishes a shared model architecture by requiring clients to agree on a set of identical layers, and leverages a public LHD dataset to facilitate global knowledge alignment across participants. During each learning iteration, clients first train their local models using private lithography data. Each trained model is then evaluated on the public dataset to generate logits. The logits, together with the parameters of the identical layers, are aggregated by the server and distributed back to the clients. Guided by the aggregated knowledge and using the public dataset as a shared medium, clients update their local models to achieve global consensus in each training round.

By jointly aggregating both model parameters and knowledge representations, FedKD-hybrid enables more comprehensive information sharing, which is particularly beneficial for learning complex multimodal patterns in LHD. We evaluate FedKD-hybrid against several state-of-the-art (SOTA) federated learning methods on the ICCAD-2012 (ICCAD) and FAB (real-world collected) datasets under various experimental settings. Experimental results consistently demonstrate that FedKD-hybrid outperforms existing approaches in terms of both accuracy and robustness for lithography hotspot detection.

In summary, the main contributions of this study include: 
\begin{itemize}
    \item We propose a hybrid knowledge distillation strategy, enabling richer knowledge transfer in collaborative learning settings. 
    \item We propose FedKD-hybrid,  a novel federated learning framework based on hybrid knowledge distillation approach, achieving enhanced learning performance in decentralized LHD learning tasks.
    \item We conduct extensive experiments on benchmark and real-world datasets to validate the effectiveness and superiority of FedKD-hybrid.
\end{itemize}

\section{Background, Motivation, and Related Work}
\subsection{Background}
As one of the most representative aggregation methods, FedAvg~\cite{2016Communication} shows effectiveness when the learning settings are close to the ideal state, i.e., Independent and Identically Distributed(IID) overall data distribution and benign learning environments. Considering $N$ clients participate in learning tasks, the optimal model $w^*$ under FedAvg is the solution for the following optimization problem, where $\ell$ is the loss function:

\begin{equation}
   w^*=\mathop{argmin}\limits_{w} \ell \left( w \right) =\frac{1}{N}\sum_{i=1}^N{\ell \left( w_i \right)}.
\end{equation}

Despite the promise, FedAvg faces a main limitation that all clients should own identical model architecture to support parameter aggregation and achieve knowledge transfer. Given the clients' heterogeneity, FL participants usually have different models based on their computing resources and storage spaces, making FedAvg challenging in real-world deployment.

Motivated to transfer knowledge in heterogeneous settings, FedMD \cite{li2019fedmd} has been proposed. Different from the FedAvg rely on parameter sharing for aggregation, each client of FedMD contributes a part of private data to form a public dataset to achieve knowledge sharing. Specifically, participant $i$ first evaluate their models on the public (contributed) dataset $D$ and submit the logits $f_i(D)$ to the server, where the logits are the prediction outcome before being transferred by Softmax. The server subsequently aggregates the logits through the following. Here, $\frac{1}{M_i}$ controls the weight across different clients $i$.

\begin{equation}
\bar{f}\left( \mathcal{D} \right) =\sum_{i=1}^N\frac{1}{M_i}{f_i\left( \mathcal{D} \right)}.
\end{equation}

Each FedMD client will download the aggregated logits and achieve the same logits on the local model through model distillation on the public dataset. Finally, participants train their models on the private datasets to start the next learning iteration.  FedMD relies on a public dataset, which may  lead to suboptimal performance if the dataset does not match the clients' local data distributions.

To address the common challenge of data heterogeneity in federated learning, the HFL-LA \cite{pan2023lithography} algorithm has been proposed. In the HFL-LA framework, the model that each client trains and utilizes is divided into two sub-models: a global sub-model and a local sub-model. The global sub-model, which is obtained from the server, is shared across all clients to consolidate common knowledge, while the local sub-model remains within each client to adapt to the non-IID local data, which can vary significantly between clients. The objective function for optimization in HFL-LA is defined as:

\begin{equation}
\mathop {\min} \limits_{w_{g},w_{l}} \left\{ F\left( w_{g}, w_{l} \right) \triangleq \sum_{i=1}^N p_i F_i\left( w_{g}, w_{l,i} \right) \right\},
\end{equation}

\noindent where $w_{g}$ represents the global sub-model parameters shared among clients. For the local data at client $i$, $F_i(\cdot)$ is the local (potentially non-convex) loss function. The parameter $w_{l} := \left[ w_{l,1}, \cdots, w_{l,N} \right]$ is a matrix, where the $i$-th column, $w_{l,i}$, denotes the local sub-model parameters for the $i$-th client.  $p_i \geqslant 0$ with $\sum_{i=1}^N p_i = 1$ represents the contribution ratio of each client. 
The key idea of this algorithm is that the global parameters are responsible for learning the shared global features, while the local parameters focus on learning the personalized local features. HFL-LA requires maintaining separate global and local sub-models for each client, which increases computational complexity and communication overhead. Additionally, the approach may struggle with scalability when the number of clients or data heterogeneity is large.

The FedGKD \cite{FedGKD} algorithm introduces a novel framework that utilizes global knowledge distillation to enhance federated learning in heterogeneous environments. Considering $N$ clients participate in learning tasks, the loss function of FedGKD is defined as follows: 

\begin{equation}
    \mathcal{L} = \mathcal{L}_{local} + \lambda \mathcal{L}_{distill},
\end{equation}

\noindent where $\mathcal{L}$ is the total loss function, representing the overall objective of model training.
 $\mathcal{L}_{local}$ is the local loss, which measures the performance of the local model on its own data.
$\mathcal{L}_{distill}$ is the distillation loss, which quantifies the knowledge transfer from local models to the global model.     
$\lambda$ is a hyperparameter that balances the contribution of local loss and distillation loss, allowing the model to adaptively learn from both local data and the collective knowledge of all devices. FedGKD can combine local learning and knowledge distillation, enabling robust model training in heterogeneous federated learning settings. However, FedGKD's reliance on knowledge distillation may increase communication overhead, and it may struggle with scalability when client heterogeneity is extreme or data privacy concerns arise.

\subsection{Motivation}
FedAvg and FedMD represent two distinct aggregation strategies for knowledge transfer across clients in federated learning: FedAvg utilizes a parameter-based approach, while FedMD adopts a non-parameter-based strategy through knowledge distillation. In the context of lithography hotspot detection (LHD), clients may use different model architectures depending on their available computing resources. Additionally, due to bandwidth heterogeneity, clients are assumed to participate in the learning process asynchronously.

While FedMD presents a potential solution for addressing these challenges and enabling collaborative learning in LHD scenarios, it is important to note its limitations. Specifically, non-parameter-based knowledge distillation may not fully leverage the knowledge learned by each local model or effectively achieve global consensus. Furthermore, local models may have some identical layers (e.g., the initial convolutional layers or fully connection layer under LHD), which non-parameter approaches might not exploit effectively. Therefore, the full potential of KD in LHD still warrants further investigation.

\subsection{Related Work}
Lithography hotspot detection has evolved significantly with the adoption of deep learning (DL) techniques, which have surpassed traditional methods like lithography simulation and pattern matching in terms of accuracy and scalability. Convolutional neural networks (CNNs) are among the most widely used architectures in this domain, with models like GoogLeNet and ResNet demonstrating excellent performance in feature extraction and hotspot identification.  
To address the issue of data imbalance, generative adversarial networks (GANs) have been employed to generate synthetic data, significantly improving model robustness by creating high-quality auxiliary training samples. These advancements highlight the capability of DL models in detecting lithography hotspots, yet they rely heavily on centralized data collection, which introduces concerns over privacy and data security \cite{chen2025trustworthy}.


Federated learning (FL) has emerged as a promising paradigm to address these challenges by enabling multiple parties to collaboratively train a global model without sharing raw data. In the context of lithography hotspot detection, FL has been successfully applied to leverage distributed datasets while preserving data privacy.
A representative work introduces a heterogeneous FL framework that combines global sub-models to capture shared patterns with local sub-models tailored to individual client datasets \cite{2021Lithography}. By explicitly separating shared and personalized components, this approach effectively mitigates the non-independent and identically distributed (non-IID) data issue that commonly arises in federated settings.



Knowledge distillation has been extensively explored to enhance the adaptability of federated learning models to heterogeneous client data distributions \cite{wu2022communication}.
For instance, FedRIR \cite{huang2025fedrir} disentangles client-specific and globally shared representations through masked client-specific learning and information distillation, thereby improving both personalization and global generalization across heterogeneous clients.

\section{Methodology}
\subsection{Overview}
To enhance the LHD performance and leverage the full potentiality of knowledge distillation, we propose a novel FedKD-hybrid framework. FedKD-hybrid leverages the advantages of both model parameter aggregation and distillation, enabling better knowledge transfer while maintaining lightweight data transmission. Specifically, we consider multiple lithography manufacturers jointly training a LHD learning task, with their lithography images remaining private and unable to leave their local devices. Due to the resource heterogeneity, these participants may have different model architectures (but include identical layers) and asynchronously join the learning task. Furthermore, a public LHD dataset is available for all participants to achieve knowledge transfer and global consensus.
\begin{figure*}   
    \centering
    \includegraphics[width=1.0\linewidth]{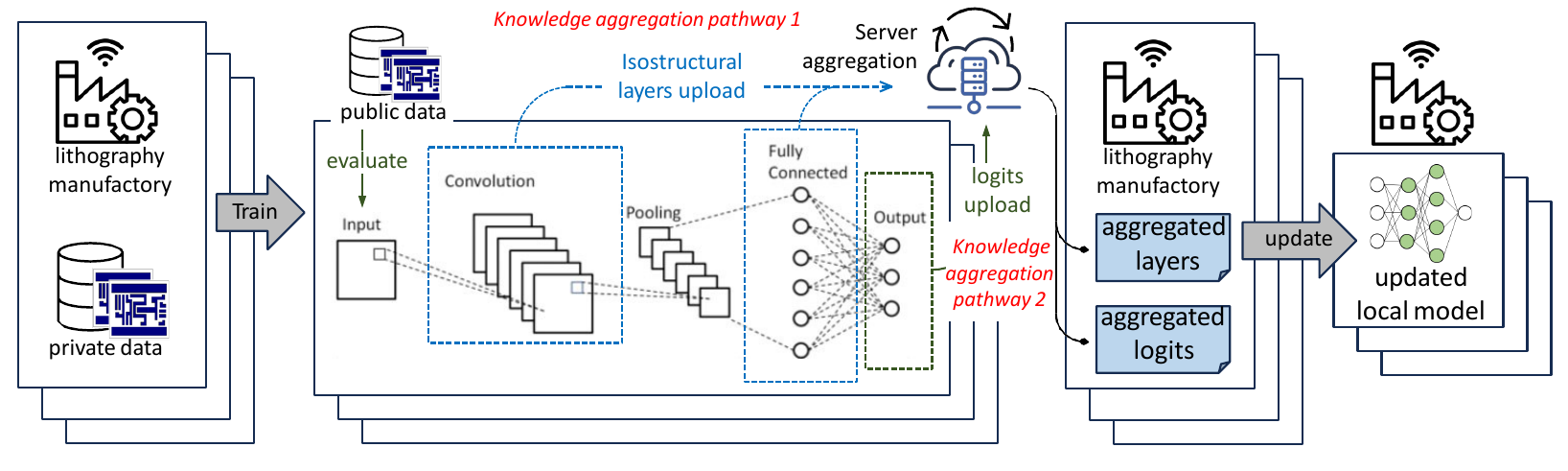}
    \caption{The overview of our proposed FedKD-hybrid algorithm.}
    \label{overview}
\end{figure*}

In the proposed FedKD-hybrid, the locally trained models are first evaluated using a public dataset to generate logits. These logits, along with selected layer parameters, are then uploaded to the server for aggregation. Participants download the aggregated information and, using the public dataset as a medium, achieve global consensus for the current learning round. As part of the knowledge gained from parameter-based aggregation, FedKD-hybrid benefits from richer information compared to traditional KD\cite{li2025frequency}, resulting in improved learning performance.

\subsection{FedKD-hybrid}
Our FedKD-hybrid includes 3 steps in each learning iteration; its overview is shown in Figure  \ref{overview}.

\textbf{Preparation}: Each client (i.e., lithography manufacturer) first agrees on the target (identical) layers to be shared during the learning task. This agreement can be reached through pre-communication or by the server specifying the expected layer architectures when releasing the learning task. To maintain the flexibility of FedKD-hybrid and ensure lightweight transmission, the server in this study sets only the first convolutional layer and the fully connected layer as identical across all participants. Additional identical layers can be introduced depending on different settings.

\textbf{Step I. Local training and logits generation}: Participants locally train their heterogeneous models using their private data, enabling effective information extraction. Once the preset training time is reached, the trained models are locally evaluated using a public LHD dataset to generate corresponding logits. In this study, we use the ICCAD-2012 dataset (an open-source LHD dataset with labels) as the public dataset $\mathcal{D}$, though it can be replaced based on privacy agreements across the lithography industry. At the end of \textit{Step I}, clients upload their upload target layers along with logits to the server. 

\textbf{Step II. Aggregation}: As shown in \textit{Algorithm 1}, once the model layers and logits from participants are received, the server performs aggregation through rule $\mathcal{A}(\cdot)$. Here, we follow the FedAvg and leverage the averaging for aggregation. The aggregated layer parameters $\bar{w}$ and logits $\bar{f}$ will be consequently sent back to all clients.

\textbf{Step III. Local model update/Knowledge transfer}: Based on the aggregated information received, clients update their local models to achieve global consensus. Specifically, the parameters of the target layers are first replaced with the aggregated values. Then, the local models are re-trained on the public dataset, using the aggregated logits as the objective. Formally, the loss function of the FedKD-hybrid is expressed as follows when averaging acts as the aggregation rule.
Here $w_{i}^{p}$ denotes the heterogeneous model layers of client $i$. $f_i=( {w_i}, w_{i}^{p})$  denotes the model of the $i$-th client, where ${w}_i$ represents the parameters of the identical model for the $i$-th client.   $
\bar{w}
$ $=$ $
\frac{1}{N}\sum_{i=1}^N{ {w_i}}
$  denotes  the aggregated parameters of the identical models across all clients.   $\lambda$ balances the contribution between the parameter and non-parameter-based knowledge transfer. Without further notification, $\lambda_i=0.5$ has been used in this study.
\begin{equation}
    F=\frac{1}{N}\sum_{i=1}^N{\left[ \ell \left( \bar{w},w_{i}^{p} \right) +\left. \lambda _i\left\| f_i\left( \mathcal{D} \right) -\bar{f}\left( \mathcal{D} \right) \right\| \right] \right.}.
\end{equation}
After completing this step, participants achieve global consensus for the current learning round, with the public dataset serving as the medium. Once \textit{Step III} is finished, clients start training the updated model on their models to start the next learning iteration.

Since the aggregated layer parameters incorporate the information learned during local training, they provide better initialization for local model updates using public data and logits, resulting in improved knowledge transfer and learning performance.

\section{Evaluation}

\subsection{Experimental Setup}
To demonstrate the performance of our proposed FedKD-hybrid, we perform comprehensive experimental evaluations under different settings. Two datasets have been used for evaluation, including ICCAD-2012 (public dataset for competition) and FAB (self-collected from industry). Specifically, we used 50\% of the ICCAD-2012 dataset to construct the public dataset for the experiments. 
We consider 100 lithography manufacturers participating in the 20-round learning task, with each maintaining a CNN model.
We use Adam optimizer 
with an initial learning rate of \textit{0.001}, a batch size of $64$ and  \textit{L2} regularization parameter   $\lambda=0.5$. Furthermore, in each round, each client performs updates for $200$ iterations on the public dataset and $100$ iterations on the private dataset.

We select several representative and state-of-the-art (SOTA) FL algorithms as benchmarks, including FedAvg ~\cite{2016Communication}, Local Training, FedProx \cite{li2020federated}, FedMD \cite{li2019fedmd}, FedPer\cite{2022FederatedPer},  HFL-LA \cite{pan2023lithography}  and  FedGKD \cite{FedGKD}. Here, FedAvg, FedProx, FedPer, HFL-LA and FedGKD are parameter-based FL aggregation methods that focus on general ML and LHD tasks, respectively. FedMD is a non-parameter-based method that utilizes knowledge distillation for model aggregation. For performance evaluation, we employ three key metrics: true positive rate (TPR), false positive rate (FPR), and accuracy.

\subsection{Experimental Results}
The experimental results show that our proposed FedKD-hybrid achieves SOTA learning performance in LHD tasks, compared with existing FL algorithms. Table ~\ref{tabI} compares the learning performance of FedKD-hybrid with different FL algorithms in ICCAD and FAB datasets in the synchronous settings. Specifically, in synchronous scenarios where all clients participate in each iteration of the learning task, our FedKD-hybrid achieves the highest testing accuracy, with scores of 0.95 on ICCAD and 0.88 on FAB. Additionally, FedKD-hybrid also demonstrates superior performance from both TPR and FPR perspectives, achieving scores of 0.99 and 0.04 on ICCAD, and 0.93 and 0.11 on FAB, respectively. In contrast, FedAvg and FedProx exhibit the lowest testing accuracies, with scores of 0.90 and 0.87 on ICCAD, and 0.55 and 0.77 on FAB, respectively. HFL-LA leverages a feature selection scheme and achieves good performance, but still lags behind our method by approximately 2\% points in accuracy.

\begin{table}[htbp]
  \centering
  \caption{Experimental performance comparison among FedKD-hybrid and the existing FL methods under synchronous settings.}
  \label{tabI}
\begin{tabular}{c|cccc}
\hline
 Dataset                & Baseline              & Accuracy        & TPR             & FPR             \\ \hline
 \multirow{5}{*}{ICCAD} & LocalTraining         & 0.9156          & 0.9869          & 0.0857
\\
                        & FedAvg                & 0.9097          & 0.9294          & 0.0907
\\
                        & FedProx               & 0.8748& 0.9343& 0.1263
\\
                        & FedMD                 & 0.9422& 0.9865          & 0.0586
\\
 & FedPer& 0.9286& 0.9738&0.0722
\\
 & HFL-LA& 0.9239& 0.9883&0.0773
\\
 & FedGKD& 0.9093& 0.9750&0.0919\\
                        & \textbf{FedKD-hybrid} & \textbf{0.9567} & \textbf{0.9901} & \textbf{0.0439}\\  \hline 
\multirow{5}{*}{FAB}   & LocalTraining         & 0.8231          & 0.8591          & 0.1786
\\
                        & FedAvg                & 0.5584          & 0.5802& 0.4426
\\
                        & FedProx               & 0.7708& 0.6945          & 0.2257
\\
                       & FedMD                 & 0.8698          & 0.8448          & 0.1290
\\
 & FedPer& 0.7826& 0.7757&0.2171
\\
 & HFL-LA& 0.8704 & 0.8615&0.1292
\\
 & FedGKD& 0.8531& 0.8063&0.1447\\
                        & \textbf{FedKD-hybrid} & \textbf{0.8889} & \textbf{0.9396} & \textbf{0.1131}\\ \hline
\end{tabular}
\end{table}

Table \ref{tabII} compares the learning performance of the FedKD-hybrid with different FL algorithms in the ICCAD and FAB datasets in the asynchronous settings. Under the asynchronous settings, where 80\% of clients randomly participate in the learning task each round, the proposed FedKD-hybrid also maintains the highest learning performance. Specifically, under the ICCAD dataset, FedKD-hybrid archives 0.95 testing accuracy, which is followed by HFL-LA as 0.94, FedAvg as 0.89 and FedProx as 0.87. HFL-LA achieves maintains the 1.6\% behind under ICCAD, but suffers from significant decrease in FAB task, receiving 76\% testing accuracy. Despite the complexity of the FAB task, FedKD-hybrid still achieves a testing accuracy of 0.88, outperforming other algorithms. From the perspective of FPR, our method receives the lowest value in both two datasets (0.044 and 0.112), demonstrating its reliability.

\begin{table}[htbp]
  \centering
  \caption{Experimental performance comparison among FedKD-hybrid and the existing FL methods under asynchronous settings.}
  \label{tabII}
\begin{tabular}{c|cccc}
\hline
Dataset                & Baseline              & Accuracy        & TPR             & FPR             \\ \hline
\multirow{4}{*}{ICCAD} & FedAvg                & 0.8954          & 0.9809          & 0.1062
\\
& FedProx               & 0.8781& 0.8592& 0.1216
\\
 & FedMD                 & 0.9289& 0.9885          &0.0722
\\
& FedPer& 0.9103& 0.9435& 0.0903
\\
 & HFL-LA& 0.9402& 0.9811&0.0605
\\
 & FedGKD& 0.8749& 0.8194&0.1241\\
& \textbf{FedKD-hybrid} & \textbf{0.9564}& \textbf{0.9896} & \textbf{0.0442}\\\hline 
\multirow{4}{*}{FAB}   & FedAvg                & 0.6091          & 0.6916          & 0.3947
\\
& FedProx               & 0.7765& 0.7124& 0.2205
\\
& FedMD                 & 0.8277          & 0.8556          & 0.1736
\\
 & FedPer& 0.7142& 0.7596&0.2879
\\
 & HFL-LA   & 0.7614    & 0.7829     &0.2396
\\
 & FedGKD& 0.7681& 0.7416&0.2307\\
& \textbf{FedKD-hybrid} & \textbf{0.8881} & \textbf{0.9071} & \textbf{0.1128} \\ \hline
\end{tabular}
\end{table}



\begin{figure}[htbp]   
    \centering
    \includegraphics[width=3.3in]{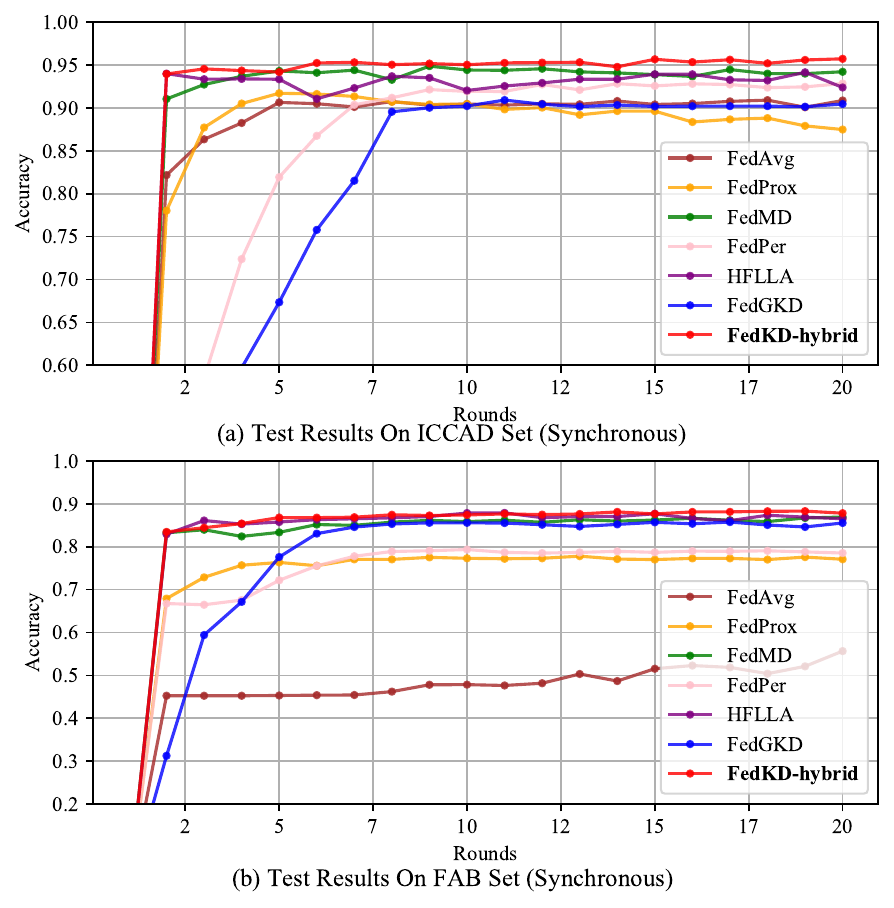}
    \caption{Test results on ICCAD and FAB set using synchronous updates.}
    \label{Plot_100Syn_FedMD_Avg_Local}
\end{figure}

We compared the performance variations of the model across various algorithms in a synchronous update scenario with 100 clients over 20 training round, where the synchronization step is set to choose all clients for training, updating, and aggregation in each round. Since only federated learning-based methods require model aggregation, we focused on comparing the federated learning algorithms in terms of their experimental accuracy during the training process. The experimental results are shown in Figure \ref{Plot_100Syn_FedMD_Avg_Local}. Compared to other state-of-the-art  federated learning algorithms, FedKD-hybrid enhances model accuracy by $1.45\%$ to $8.19\%$ on the ICCAD clients. For the FAB clients, FedKD-hybrid improves model accuracy by $1.85\%$ to $33.05\%$.

\begin{figure}[htbp]
    \centering
    \includegraphics[width=3.3in]{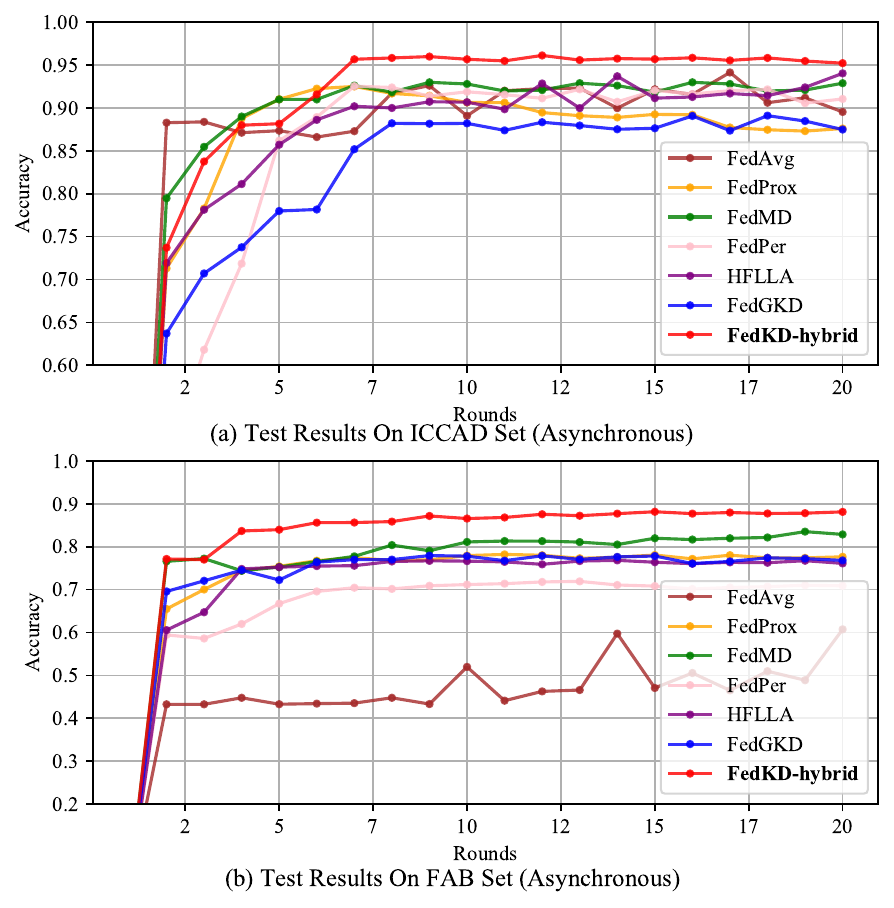}
    \caption{Test results on ICCAD and FAB set using 80\% asynchronous updates.}
    \label{Plot_80Asyn_FedMD_Avg_Local}
\end{figure}



We also compared the model performance   of various algorithms in an asynchronous update scenario with 100 clients over 20 training rounds , where the asynchronous step is set to randomly select 80\% of the clients for training, updating, and aggregation in each round. The experimental results are shown in Figure \ref{Plot_80Asyn_FedMD_Avg_Local}. As observed, in the asynchronous scenario, the model fluctuates significantly in the early stages of training due to insufficient aggregation. Among the federated learning algorithms, even with asynchronous updates, FedKD-hybrid achieves the highest accuracy and converges much faster than the other federated learning algorithms. Compared to the FedAvg algorithm, the convergence speed of FedKD-hybrid improves by at least 4 times. Compared to other state-of-the-art  federated learning algorithms, FedKD-hybrid enhances model accuracy by $1.62\%$ to $8.15\%$ on the ICCAD clients. For the FAB clients, FedKD-hybrid improves model accuracy by $6.04\%$ to $27.91\%$.

Thanks to the novel knowledge transfer, our FedKD-hybrid benefits from richer information compared to aggregation methods based solely on parameters (e.g., FedAvg) or non-parameters (e.g., FedMD), resulting in enhanced learning performance. Furthermore, since only Conv1 and FC2/3 layers are transmitted along with logits, the communication cost of FedKD-hybrid is comparable to non-parameter-based FL aggregation methods and significantly lower than that of parameter-based approaches like FedAvg.

\section{Conclusion}
In this paper, we have proposed the FedKD-hybrid algorithm to enable collaborative training in privacy-preserving LHD learning tasks. In FedKD-hybrid, a public dataset is used to achieve global consensus, and several layers are designated as identical. The trained identical layers and logits from evaluations on the public dataset are aggregated by the server and then broadcast to support local updates. By leveraging both parameter and non-parameter knowledge transfer, FedKD-hybrid is expected to achieve superior learning results. We evaluate FedKD-hybrid on two datasets under different settings, and the experimental results demonstrate that FedKD-hybrid delivers SOTA performance compared to existing FL methods while maintaining low communication costs. 

{
\small 
\bibliographystyle{ieeebib}
\bibliography{references}
}

\end{document}